\def\year{2022}\relax
\documentclass[letterpaper]{article} 
\usepackage{aaai22}  
\usepackage{times}  
\usepackage{helvet}  
\usepackage{courier}  
\usepackage[hyphens]{url}  
\usepackage{graphicx} 
\usepackage{natbib}  
\usepackage{caption} 
\DeclareCaptionStyle{ruled}{labelfont=normalfont,labelsep=colon,strut=off} 
\usepackage{algorithm}
\usepackage{algorithmic}
\usepackage{multirow}
\usepackage{makecell}

\usepackage{xcolor}         

\usepackage{newfloat}
\usepackage{listings}
\floatstyle{ruled}
\newfloat{listing}{tb}{lst}{}
\floatname{listing}{Listing}
\newcommand{\sysname}{{\sc Creative-Wand}}

\newcommand\Mark[1]{}
\newcommand\MarkLeft[1]{}
\newcommand\MarkRight[1]{}
\newcommand\Zhiyu[1]{}
\newcommand\ZhiyuLeft[1]{}
\newcommand\ZhiyuRight[1]{}

\title{Creative Wand: A System to Study Effects of Communications in Co-Creative Settings }
\author{
    Zhiyu Lin,
    Rohan Agarwal,
    Mark Riedl
}
\affiliations{
    Georgia Institute of Technology

    North Ave Northwest, Atlanta, Georgia 30332, United States of America

    \{zhiyulin,roaga\}@gatech.edu, riedl@cc.gatech.edu

}

\usepackage{bibentry}

\usepackage{booktabs}

\begin{document}

\maketitle

\begin{abstract}
Recent neural generation systems have demonstrated the potential for procedurally generating game content, images, stories, and more.
However, most neural generation algorithms are ``uncontrolled'' in the sense that the user has little say in creative decisions beyond the initial prompt specification.
Co-creative, mixed-initiative systems require user-centric means of influencing the algorithm, especially when users are unlikely to have machine learning expertise.
%
The key to co-creative systems is the ability to {\em communicate} ideas and intent from the user to the agent, as well as from the agent to the user.
Key questions in co-creative AI include:
How can users express their creative intentions?
How can creative AI systems communicate their beliefs, explain their moves, or instruct users to act on their behalf?
When should creative AI systems take initiative?
The answer to such questions and more will enable us to develop better co-creative systems that make humans more capable of expressing their creative intents.
We introduce \sysname{}, a customizable framework for investigating co-creative mixed-initiative  generation.
\sysname{} enables plug-and-play injection of generative models and human-agent communication channels into a chat-based interface.
It provides a number of dimensions along which an AI generator and humans can communicate during the co-creative process.
We illustrate the \sysname{} framework by using it to study one dimension of co-creative communication---global versus local creative intent specification by the user---in the context of  storytelling.

\end{abstract}

\begin{figure*}
    \centering
    \includegraphics[width=\linewidth]{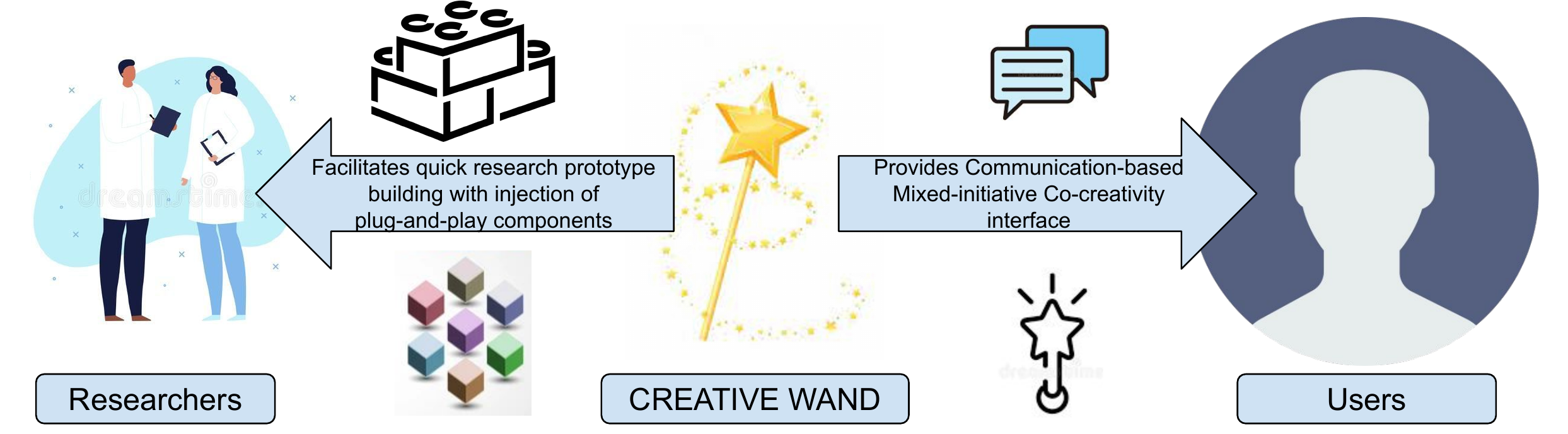}
    \caption{\sysname{}  is a customizable framework that helps researchers quickly iterate on prototypes of mixed-initiative co-creativity applications with Application Interfaces defined for components ranging from generative systems to AI mediators and interfaces, based on our Communication Model.}
    \label{fig:teaser}
\end{figure*}

\section{Introduction}

Generative AI systems can support artists, writers, and all sorts of creative work.
{\em Neural generative systems} \cite{khalifa_pcgrl_2020,brown_language_2020,liu_pre-train_2021,radford_learning_2021,creswell_generative_2018,ho_denoising_2020,ramesh_hierarchical_2022}
%
have increasingly been used by non-technical artists, writers, and others, whom we collectively refer to as {\em human designers} for the context of this paper.
Originally, human designers had to collect and curate datasets to train neural generative systems.
Increasingly, the availability of large pretrained neural generative models means human designers can condition a model on a prompt \cite{liu_pre-train_2021}, previous context \cite{brown_language_2020}, structured data \cite{lebret_neural_2016}, or even multi-modal inputs \cite{radford_learning_2021} in order to exert influence over a model's outputs.
However, these means of data input, along with most interactions in and out of these generation systems, are not human-centered in the sense that they impose paradigms of interaction on the human designer that is afforded by the underlying algorithms and models instead of what best suits the needs of the human designer. 
This can result in greater cognitive load, frustration, and ultimately reduced use of a system
\cite{sweller_cognitive_2011}.

%
One solution to this problem is to introduce machine co-creativity and mixed-initiativeness~\cite{liapis_can_2016}
so that AI and users work together.
\citet{kreminski_reflective_2021} note there is a wide range of types of interactions that can happen between the AI agent and the human designer.
The decision on what type of interactions to allow and when to deliver them 
has implications on whether a human designer succeeds in achieving their creative intent.

At the heart of co-creative agents is {\em communication}; the human designer must convey their goals, intentions, and desires so that the agent can act upon them to the benefit of the user. 
The agent may also need to ask for clarification or assert its own intentions back to the human designer. 
This communication can be explicit or implicit; it does not necessarily need to be in natural language.
However,
the communication between human designers and co-creative agents is often overlooked as critical to the success of a co-creative system.
To understand the possible solutions to how human designers and AI communicate and how they impact the human-AI co-creative teaming, we must first understand the space of different co-creative design solutions when it comes to when and how the user and agent communicate.


We introduce \sysname{},\footnote{Available at https://github.com/eilab-gt/CreativeWand} a customizable framework for investigating co-creative mixed-initiative text generation systems.
\sysname{} enables plug-and-play injection of neural {\em generators}, 
human-agent \textit{communication types} and visual interfaces into a co-creativity, mediated by an \textit{Experience Manager}.
We also provide an ontology of ways in which humans and AI generators can communicate, broken into three dimensions: human-initiated vs agent-initiated, elaboration vs. reflection, and global vs local.
We illustrate the \sysname{} framework with a study of one dimension of co-creative communication: global versus local intent specification by the user in the context of neural story generation.

Our contributions are as follows.
    (1)~We formalize dimensions of communication between the agent and the user.
    (2)~We report on \sysname, a system to facilitate rapid iterations of research on the design space of user-agent communications in co-creative mixed-initiative settings.
    (3)~We demonstrate how \sysname{} can be used to study co-creative systems by conducting a study of agent-user communication along one dimension.

\begin{figure}[t]
    \centering
    \includegraphics[width=\linewidth]{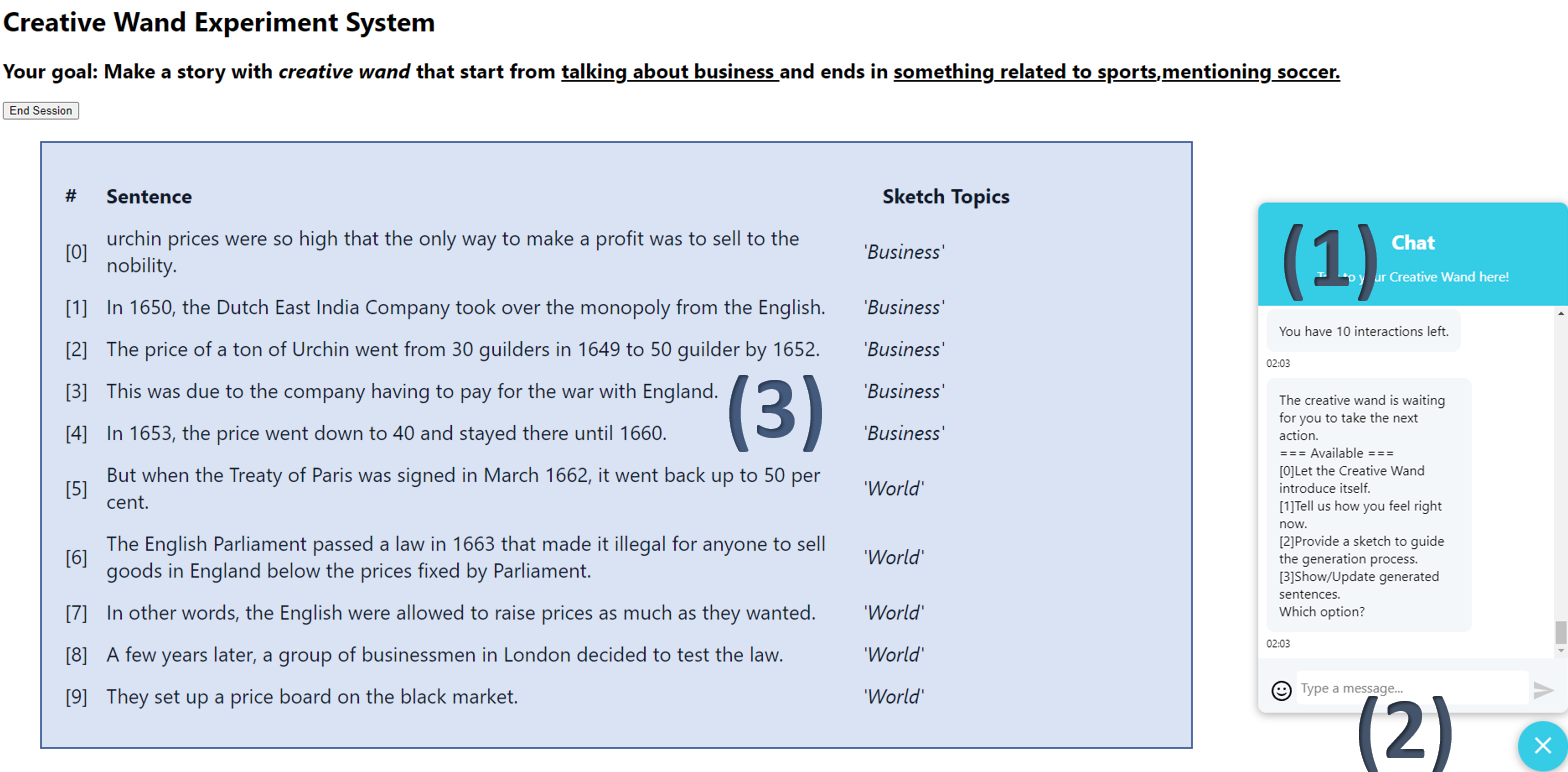}
    \caption{A screenshot of the instantiation of \sysname{} used in the experiment. 
    }
    \label{fig:screenshot}
\end{figure}

\section{Background and Related Work}

A mixed-initiative system is one where ``a human initiative and a computational initiative" cooperate towards a shared goal~\cite{novick_what_1997}.
Mixed-initiative systems can be applied to the problem of producing content for computer games~\cite{yannakakis_mixed-initiative_2014}, or, more broadly, creative works.
We refer to an intelligent system as a {\em co-creative} agent\Mark{is this consistent?} when it possesses the ability to alter the creative work equal to a human counterpart.
Note that ability does not imply human-parity when it comes to capability.
Ability also does not imply ``responsibility'' as the human and the AI system may assume responsibilities for different aspects of the creative artifact or the creative process.

Multiple researchers have attempted to categorize or differentiate between different types of interactions between the creative agent and the users.
\citet{rezwana_cofi_2021,rezwana_designing_2022} proposed a framework that models interactions in co-creative systems with a focus on the actual presentation and flow of information between collaborators.
\citet{guzdial_interaction_2019} identified a general framework of human and AI exchanging ``artifact'' and ``other'' actions in an implied turn-taking fashion, but also allowing for non-turn actions.
\citet{grabe_towards_2022} described a Generative-Adversarial-Network (GAN)-based framework.
Both Guzdial and Grabe point out that human designers and AI can initiate the \textit{same} ``action'' sets to modify the creative work, albeit with different executions, inspiring the human-vs-agent-initiated dimension in our ontology of communication types.
\citet{kreminski_reflective_2021} conducted a survey on a subset of interactions, specifically agent-initiated reflective ones, where the agent thinks about what happened in the process and takes actions based on it, inspiring the elaboration vs. reflection dimension in our model.
We attempt to refine our understanding of how humans and AI systems communicate during co-creative design; our ontology of communication augments these existing frameworks.

Deep neural networks for image generation have become increasingly prevalent, using Generative Adversarial Networks (GANs) or diffusion techniques. 
Techniques such as DALL-E \cite{ramesh_zero-shot_2021,ramesh_hierarchical_2022} and Imagen~\cite{saharia_photorealistic_2022} use text-to-image classifiers such as CLIP~\cite{radford_learning_2021} to receive a textual prompt and guide the generative process until it matches the prompt to some degree. 
Most of this work is not co-creative in the sense that the human user only provides the initial prompt and has no further ability to influence the generation process.\footnote{Aside from using a different prompt and restarting the process.}
Some GANs have been incorporated into interfaces, like ArtBreeder~(\url{https://www.artbreeder.com/}) to allow the user to browse the latent space of the model, searching for images that suit their creative goals.\Mark{There was also something in CHI. See inline.}
%
Deep neural networks have also been applied to writing assistance, initially with LSTMs~\cite{roemmele_writing_2016} and later with transformer-based language models \cite{thoppilan_lamda_2022,wolf_huggingfaces_2019}.
In most cases, users provide an initial sentence as a prompt, and the language model generates a continuation of the text. 
Writing assistance tools built around language models may allow the user to go back and manually edit, or re-generate portions of text.
The LaMDA transformer~\cite{thoppilan_lamda_2022} is notable in that it supports creative writing by providing a number of different ways of prompting the system at different stages of writing.

While most text generation with transformer-based language models starts with a user-provided first sentence as a prompt, a number of research efforts in story generation attempt to modify the AI system to allow for a greater degree of control over what is generated.
One means of controlling story generation is to condition generation on high-level plot outlines~\cite{fan_hierarchical_2018,peng_towards_2018,rashkin_plotmachines_2020}
or story in-filling~\cite{donahue_enabling_2020,ammanabrolu_bringing_2020}\Mark{durrett story infilling paper see inline comment for bib}.
\citet{tambwekar_controllable_2019} and \citet{alabdulkarim_goal-directed_2021} retrain the language with a specific, provided end-goal. 

Our experiments use the {\em Plug and Blend}~\cite{lin_plug-and-blend_2021} controllable generation technique, which trains a neural network to modify the logits produced by a neural language model in order to induce the generation of text that conforms to a given topic. 
Plug and Blend can further receive multiple topics and blend those topics together to generate text that smoothly transitions from one topic to another.
Users can theoretically provide an initial sentence prompt as well as a schedule of topics called a ``sketch''.
While the original Plug and Blend operated with the GPT-2 language model, we have updated it to GPT-J~\cite{wang_gpt-j-6b_2021}, a 6B parameter language model.

\begin{figure*}[ht]
    \centering
    \includegraphics[width=\linewidth]{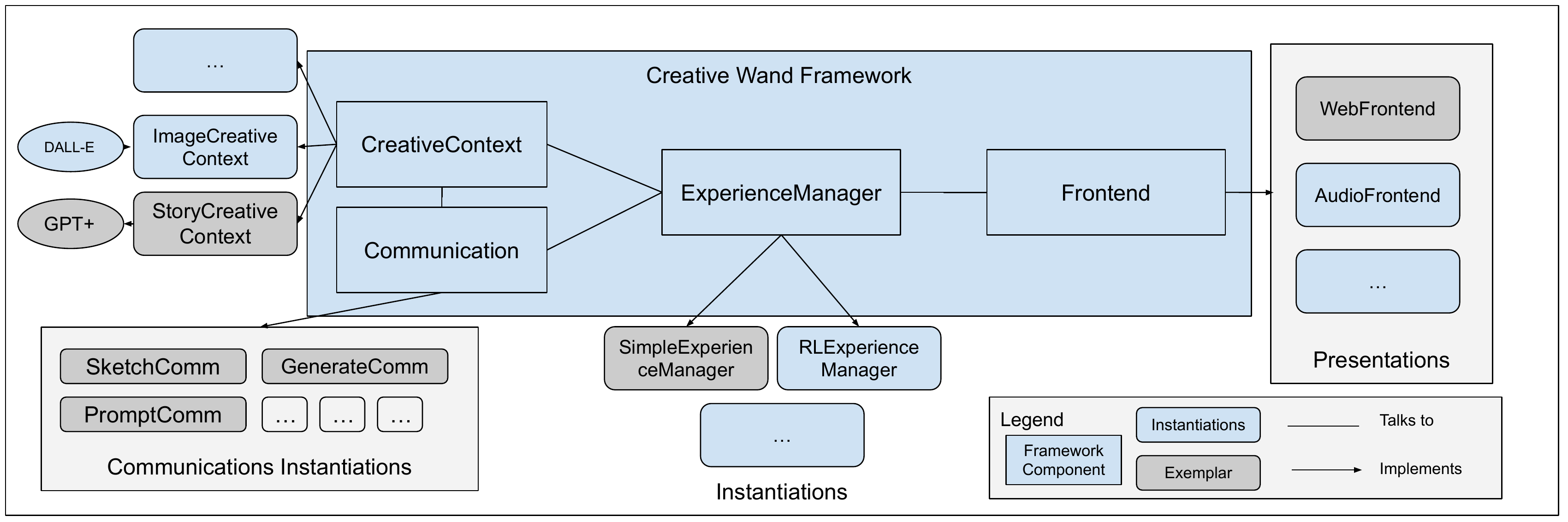}
    \caption{\sysname{}, the plug-and-play mixed-initiative co-creativity framework, with potential instantiations. Instantiations used in the study are labeled gray.
    \Mark{if you have time, we can save a lot of space by making this image tall instead of long and putting it in one column}
    }\Zhiyu{Hmm. I will have to think about a clever layout for portrait mode of this. If we ended up needing more space. }
    \label{fig:framework}
\end{figure*}

\section{Co-Creative Communication}
\label{section:paradigm}

Every creator uses tools in the creative process.
Tools are artifacts that augment or extend the user's abilities. 
For example, a hammer enhances one's striking power.
A paintbrush enhances one's ability to apply paint to a medium consistently.
We often think of tools as simple, but they can be complex and even contain degrees of intelligence. 
A word processor enhances one's ability to revise text and includes some degree of initiative to identify spelling and grammar mistakes, and agency with regard to how to correct them.
A generative network for creating pictures 
can be thought of as a tool with a high degree of agency---a user enters a phrase and the tool chooses the color for every pixel in the image.

A {\em co-creative} tool may possess unique capabilities not traditionally found in conventional tools:
    {\bf (1)~Agency:} the tool has the ability to make decisions in response to high-level specifications from a user.
    {\bf (2)~Initiative:} 
    the ability to decide to alter the creative work without explicit permission from the user.
    {\bf (3)~Communication:} the ability to send and receive information (verbally or otherwise) about the creative work, creative processes, or the agent's internal state, including goals, instructions, feedback, or explanations.   

Whereas Agency is a property of intelligent tools, Initiative is a property of {\em mixed-initiative tools} that can assert agency without being directed to do so by the human designer.
{\em Communication}, on the other hand, is what makes a tool {\em co-creative}, and is often overlooked.
For a tool to be considered co-creative, it must be able to send and receive
information, such as goals, directives, intentions, or criteria to and from the user.
In this way, an intelligent tool becomes a partner to the user.
This is especially important as the tool may be asserting its agency to make decisions in response to high-level directives, and may additionally avoid failures by asking for clarifications and to arrive at a shared understanding of the goal and success criteria of the creative process;
It should be noted that the 
transfer of information need not be in natural language.



Highlighting the role of communication in co-creative agents still leaves numerous possible designs for how, when, and what humans and agents communicate.
We further refine our understanding of the co-creative agent design space by identifying three (non-exhaustive) dimensions of human-agent communication, given below.



\textbf{Human-initiated vs. Agent-initiated.}
The first dimension considers which of the two parties is initiating communication, and by implication who is the recipient.
An instance of communication can be initiated by the human designer or by the agent. 
This dimension does not consider the content or the nature of the communication but looks at the directionality of the communication.
Instead of treating a co-creative process as a process led by a certain party---for example, by the agent, where the human reacts to what the agent does---we follow the theory of mixed-initiative \cite{yannakakis_mixed-initiative_2014}
where no priority is given to any party in regard to their initiative or agency. 


\textbf{Elaboration vs. Reflection.}
This dimension deals with whether the communication relates to previously generated contents (reflection) or to newly planned actions (elaboration).
Elaboration refers to communication about future actions to revise the creative content that has not yet been performed. 
Elaboration includes: dictating goals 
(``The story needs to be about a particular topic'');
instructing the other to perform actions 
(``Explain what a character does next'');
seeking clarification about a directive 
(``what is a character feeling?'');
or providing clarification 
(``The character is afraid'').
%
Reflection~\cite{kreminski_reflective_2021} is the action of "reflecting on their own work... to surpass limitations".
Examples of reflective communication include: 
feedback about goals 
(``I didn't want the character to come to harm'');
revisions 
(``this should have been done differently'');
 explanation
(``I added this sentence because...'');
and sharing of evaluation
(``The last edit you made looks on topic.'')
%

The dimension of elaboration vs reflection shares similarities to the engagement-reflection model of creativity by~\citet{sharples_account_1996}, which describes the process of actively adding content with minimal consideration of constraints (engagement) and revisiting and revising previously added content (reflection).
However, our dimension is not about the creative processes but about the types of communication about how the content should be changed.


\textbf{Global vs. Local Scope.}
The final dimension is the scope of the creative work that is targeted by the communication. 
Communication can be targeted at different scopes of content, ranging from global to local.
{\em Global scope} indicates aspects that apply to the entirety of the creation. 
For example, in the context of storytelling, a request for a stylistic change, or leading the protagonist towards a romantic relation.
{\em Local scope} indicates an individual aspect, detail, or element, such as a request to add a tree in front of a castle. \ZhiyuLeft{Made edits here so they are all based on stories now.}
Some communications may fall in the middle, as when talking about a region or a group of entities.

\section{\sysname}
\label{section:framework}

Research questions pertaining to 
how
users can express their creative intentions, 
how creative AI systems can communicate their beliefs, explain their moves, or instruct users to act on their behalf, and
when creative AI systems should take initiative to require humans in the loop. 
There are many ways in which a creative AI system can express agency, take initiative, or communicate with users that may be independent of how the AI generative algorithm operates.
To explore the space of co-creative agent designs, we have developed \sysname{}, which allows AI generative algorithms to be situated within a human-in-the-loop interface through a well-defined application interface (API) specification.
\sysname{} supports the communication ontology presented in the previous section, enabling experimentation with different communication configurations.

Figure~\ref{fig:screenshot} shows a screenshot of the \sysname{} user interface under the setting of co-creating stories. 
To support language-based communication, it provides a short message text (chat) interface so that the agent can communicate with the user~(1) and the user can communicate back to the agent~(2) through short text messages.
If this communication channel is used, then it is up to the underlying agent to be able to generate messages and parse messages from the user.
There are no constraints on the format of the messages---the agent could implement a menu system where the user enters numbers for options, or a simple command language, all the way up to natural language.
The majority of the screen is a canvas~(3) in which to render the creative work and additional non-language communication from the agent to the user.
In the figure, we show the user and agent collaborating on a short story, so the creative work is presented as a sequence of sentences, numbered so that the user can refer to them in chat.
This particular agent (described below) can also render some of its internal data structures to assist the user.


\subsection{Architecture}

\sysname{} is a lightweight framework that provides a minimalistic plug-and-play API for AI algorithms and for communication handlers.
Figure~\ref{fig:framework} shows how modules are structured and how they interact with each other.
There are five abstract modules that must be instantiated.

    \textbf{CreativeContext.}
    This module implements the generative algorithm that receives prompts or other specifications and produces a creative artifact to the best of its ability.
    Examples include image generators, story generators, or game level generators. 
    The AI system must support two API calls: 
    \verb|execute_query()|, which specifies instructions for the  generator in a protocol that it understands, and
    \verb|get_generated_content()|, which presents generated artifacts to other parts of the system.

    \textbf{Communication.}
    This module instantiates the code that determines how information is passed back and forth between the CreativeContext and the user.
    \verb|activate()| defines the behavior of the communication, whether it's an elaboration or a reflection, and what information will be presented to and received from the FrontEnd,\footnote{Communication modules do not directly manipulate the front-end, it is left to the Frontend to determine the actual presentation.}
    and what is needed from the CreativeContext.
    \verb|confidence_to_activate()| and
    \verb|confidence_to_interrupt_activate()| are respectively used by the ExperienceManager to determine whether a human designer can activate this communication (human-initiated), and to determine whether this communication is ready to ``interrupt'' and assume initiative (agent-initiated).
    There will be multiple communication modules, one for each way the user can instruct/inform the agent and that the agent can instruct/inform the user.
    
    \textbf{ExperienceManager.}
    This module is responsible for maintaining the \verb|state| of the system. Key interfaces include 
    \verb|activate_preferred()|, which determines which communication should be initiated by the agent (or not at all), and
    \verb|interrupt_activate()|, which allow user to activate available communications (human-initiated).
    A basic implementation of the ExperienceManager would only use confidence signals from Communications made available to it.
    More complex implementations are possible, such as ones that learn communication preferences.
    
    
    \textbf{Frontend.} 
    The co-creative agent nevertheless needs to interact with the human designer.
    To adapt to a variety of presentations, 
    this module is implemented to provide whatever type of output is needed: a command line, HTML panel, drawing canvas, audio, etc.
    A Frontend is defined on how it \verb|get()| information from the user and how other components \verb|send()| information to it.

Additionally, \sysname{} provides configurable logging and tools for quantitative evaluation.
Table~\ref{table:instantiation} shows one possible instantiation of \sysname{} modules.
This particular set of instantiated modules is also used in the experiment in the next section.





\begin{table*}[htbp]
\centering
\begin{tabular}{c|p{0.22\linewidth}|p{0.53\linewidth}} 
\toprule
{\bf Module} & {\bf Instantiation}           & {\bf Description} \\ 
\hline
{\bf CreativeContext}    & StoryCreativeContext    & A backend interfacing with an implementation of Plug and Blend \cite{lin_plug-and-blend_2021} with GPT-J\cite{wang_gpt-j-6b_2021} as the base language model, supporting both prompts and "sketch-based" high-level control.  \\ 
\hline
{\bf ExperienceManager}  & SimpleExperienceManager & A turn and rule-based agent that shows all available Communications and allow the user to make a choice, or request for activation of Interrupted Communication when there is one.~                 \\ 
\hline
{\bf Frontend}           & WebFrontend             & A React.js web application with Chatbox interface and a canvas showing the artifact and additional information. (See figure \ref{fig:framework})                                                                            \\
\hline
\multirow{2}{*}{\bf Communications} & Local condition: \textit{UserWorkComm}, \textit{GenerateWithFreezeComm} & Unique to this condition, allow the user to manually edit any line in the story, and ``freeze'' any line of the story\\
\cline{2-3}
 & Global condition: 
 \textit{UserSketchComm} 
 & Unique to this condition, provides a global communication type allowing the user to set topics for a "sketch"\cite{lin_plug-and-blend_2021} to influence part of the story. Also see Figure  \ref{fig:screenshot} for an example.\\
\bottomrule
\end{tabular}
\caption{\sysname{} module instantiations used in the experiment.}
\label{table:instantiation}
\end{table*}

\subsection{Protocol and Metrics}

\sysname{} facilitates the exploration of the space of possible designs for communication between a human user and a co-creative agent. 
Given that there can be many instantiations of CreativeContext as well as Communication modules, one must have a uniform set of metrics across which to compare different co-creative instantiations.
The primary use case for \sysname{} is to keep all modules constant, such as the generator, and vary the communication types supported. 
In doing so, one can determine whether certain types of communication---as points within the space defined by our proposed communication dimensions---support or hinder human creative practice. 

For the purpose of running controlled human participant studies, we recommend a protocol in which a number of participants are asked to create an artifact that meets a given set of criteria (goals). 
While it is possible for creation to occur in a completely exploratory fashion, a significant portion of creation is in pursuit of a specific goal, which may be compound or have sub-goals. 
For example, one may have specific criteria in mind for a generated story or image or game level~\cite{riedl_lovelace_2014,cherry_quantifying_2014}.
The recommended protocol provides study participants with one or more criteria to achieve within a fixed amount of time.
The first metric is {\bf goal completion}: how many of the goals were completed in the time available.
Due to the subjective nature of whether a participant believes a goal was achieved and may defy automatic decisions, we propose to measure this through self-report and can additionally measure the number of user-agent interactions at time of report as an indicator of ease. 

Subjective dimensions include frustration and satisfaction.
{\bf Frustration} is the extent to which a participant feels frustrated with their ability to affect desired change and to achieve given criteria through the communication channels available. 
Frustration is a common experience with creative AI systems in which the user can only provide a single prompt and must repeatedly re-generate to get a desired outcome.
{\bf Satisfaction} is the extent to which a participant reports that they are satisfied with the quality of the work. 


\section{Experiments}

We demonstrate \sysname{} by conducting the protocol described above to investigate one part of the co-creative agent communication design space.
Specifically, we look at whether a version of \sysname{} that implements only global communication affects goal completion and frustration when compared to a version that allows only local communication. 
%

The \sysname{} instantiations are summarized in Table~\ref{table:instantiation}.
In both conditions, we use the Plug and Blend generator~\cite{lin_plug-and-blend_2021} applied to GPT-J~\cite{wang_gpt-j-6b_2021} because its sketch inputs provide global control of topics of sentences in a story.
We implement two versions of \sysname, one for each condition in our study.
In the {\bf global condition}, the global communication type {\em UserSketchComm} allows the user to add topics to the sketch. 
In the {\bf local condition}. participants have the ability to manually edit any line in the story, and to communicate that the agent should ``freeze'' a certain line when re-generating so it won't be overwritten.
Communication types shared by both conditions allow the user to ask the agent to re-generate the story and to indicate when the have completed a sub-goal or when they feel frustration. 

The front-end module is as shown in Figure~\ref{fig:screenshot}, and the experience manager is a simple pass-through.
Except for suggesting freezing the line that the user just manually edited in the Control Condition to facilitate the process (otherwise risking the line getting overwritten), neither the Plug and Blend + GPT-J agent nor any other models take initiative nor issue interrupts.

We follow the general protocol outlined in the previous section, making details concrete here.
The participants first answer some background questions related to their experience in computing, game design, and AI.
They will then be randomly assigned one condition and be given information related to the overall use of the system, along with a short description of key communications available to them for the condition.
They are instructed to ``Make a story with that (1)~starts from talking about business and (2)~ends in something related to sports, (3)~mentioning soccer". 
Participants then proceed to work on authoring stories with ten available lines. The three criteria (sub-goals) were carefully chosen to not be trivially achievable in either condition; the criteria that the story ``mentions soccer'' cannot be achieved by giving the agent a global topic, and requires some repeated regeneration.
Participants are allowed 15 interactions with \sysname{}, defined as triggering of any Communication other than providing feedback;
We found in internal piloting that 15 interactions would grant enough time for participants to finish the task assigned without they losing track of it.
Participants can also indicate through a special communication type that a sub-goal is completed and also ``how they feel''.

Once they used up allowed interactions or opt to end the session 
early, they are asked, on Likert scale, the extent to which they agree or disagree
((Strongly) Disagree, Neutral, (Strongly) Agree) with statements about achieving each of the three goals, feeling satisfied with their story, and feeling frustrated.

We recruited 60 participants on the crowdsourcing platform {\em Prolific}.
Participants were paid \$15 per hour.
Participants took on average 20  minutes to complete the task.


\section{Results and Discussion}
97\% of the participants report at least some experience using computer for creative work, while 40\% use them in their job.
When it comes to game development, a likely context for using creative AI tools, 46\% reported some experience while only one participant reported it as part of their job.
84\% report to at least have heard of AI and 30\% report understanding how recent AI technologies work.

Participants self-report when they complete any of the given sub-goals. 
Note that as each participant can start fresh by restarting a session, we only take sessions that have at least two interactions and only take the best out of one participant on each metric. 
Table~\ref{table:exp_completion} shows self-reported completion rates.
There is substantial under-reporting, especially on later goals.
Participants in the global condition report more goal completion ($p<0.1$)
for the first goal (beginning the story with the ``business'' topic), which is one of the sub-goals we anticipated would be facilitated by communicating sketch information.
There is, however substantial under-reporting of goals 2 and 3.

When participants report goal completion, we also record how many interactions have been completed.
Table~\ref{table:interactions} shows the interaction count at time of report for each goal. 
Consistent with the above results, participants in the global condition achieve the first goal faster ($p=0.14$).
Participants in the global condition also satisfied the second goal (ending the story with the ``sports'' topic) with significantly fewer interactions ($p<0.05$). 
This goal also aligned with global sketch communication. This result is in line with untested claims by \citet{lin_plug-and-blend_2021} about the ease of generating transitions to new topics.

Regarding frustration, 31.3\% of the global condition participants reported frustration at least once while 28.6\% did in the local condition (two-sided $p=0.82$ for $H_{0}: p_{exp} = p_{ctrl}$), which means we cannot statistically distinguish between conditions on the dimension of frustration.
Frustration may emanate from the AI generator, user interface, communications types available, or other sources.
Initial qualitative analysis of the logs suggests that those in the global condition struggled to achieve the third goal (``mention soccer'') because the only means of achieving it was through regeneration. 
Those in the local condition struggled to satisfy the first two goals and many satisfied all sub-goals by manually editing story lines.\Mark{fair?}\Zhiyu{Yep I see the same.}

\begin{table}
\centering
\footnotesize
\begin{tabular}{c|c|c|c}
{\bf Reports of sub-goal Completion}                     & {\bf \#1}    & {\bf \#2}    & {\bf \#3}     \\
\hline
Local condition (n=28)      & 25.0\% & 17.9\% & 10.7\%  \\
Global condition (n=32) & 40.6\% & 25.0\% & 15.6\%  \\
p-value ($H_{0}:p_{global}\leq p_{local}$)                      & 0.095  & 0.249  & 0.285 \\  
\hline
\end{tabular}
\caption{Metrics on rate of sub-goal reported as completed.}
\label{table:exp_completion}
\end{table}

\begin{table}
\centering
\footnotesize
\begin{tabular}{c|c|c|c}
{\bf Interactions needed for sub-goal}                & {\bf \#1} & {\bf \#2} & {\bf \#3}  \\
\hline
Local condition      & 8.71 & 9.40 & 9.33  \\
Global condition & 7.08 & 6.25 & 5.80 \\
p-value ($H_{0}:t_{global}\geq t_{local}$)               & 0.140 & 0.027 & 0.047 \\
\hline
\end{tabular}
\caption{Metrics on interactions taken to achieve sub-goals.
}
\label{table:interactions}
\end{table}

\begin{figure}
    \centering
    \includegraphics[width=\columnwidth]{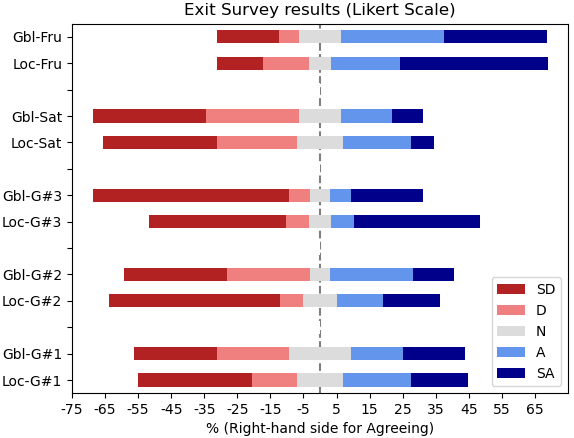}
    \caption{Results from the exit survey. Dimensions are abbreviated: 
    \textbf{Loc} for Local Condition;
    \textbf{Gbl} for Global Condition;
    \textbf{G\#x} for the x-th sub-goal completion;
    \textbf{Sat} for satisfaction; \textbf{Fru} for frustration.
    }
    \label{fig:exitsurvey}
\end{figure}

%
Figure \ref{fig:exitsurvey} shows the exit survey results.
Local condition participants can manually edit and freeze sentences and as a consequence we observe significantly more participants strongly agree ($p<0.1$) that they achieved sub-goal 3 than those in the experimental condition.

We observe substantial number of participants expressing frustration on the system in the exit survey (more than self-reporting during interaction), likely due to the fact that the exit questionnaire is required whereas reporting during creation is voluntary.
Although we focused on demonstrating the communicational \textit{capability} of the system, we acknowledge that \textit{how and when} communications should happen between both parties remains a major subject of study;
Capturing feedback is a prime example: 
On one hand, only passively capturing feedback, like what we did in our experiment, led to potential underreporting, and the AI may not have enough data to potentially improve the creative session;
On the other hand, letting AI ask for too much feedback have the potential of lengthening creative sessions, causing a heavier cognitive load, and creating annoyance.
We leave these research questions as future work and invite the community to further study adaptive, personalized co-creativity systems using \sysname.


The purpose of the experiment was primarily to demonstrate the way in which \sysname{} facilitates research on the design space of co-creative agent communication with users. 
To that end, the comparison of global-only versus local-only communication types is artificial.
Likely co-creative agent solutions will use a mix of communication strategies.
To that end, we observe participants in the global condition being frustrated by the lack of local communication and vice versa. 
However, we do show that global communication through topic sketches---a limited form of global communication in our ontology---can be beneficial for achieving certain creative criteria. \Zhiyu{Great discussion!}

\section{Conclusions}

We describe \sysname{}, a general, flexible, plug-and-play framework for studying mixed-initiative co-creativity.
We present an ontology of co-creative \textit{communications} between the human designer and the agent and use it to demonstrate how specific instantiations of abstract components with \sysname{} can support experimentation to understand the space of designs of user-agent interaction that might be possible in a mixed-initiative co-creative setting.
We further report the results of a study conducted with an exemplar instantiation of \sysname{} for story generation that shows the importance of being able to communicate about global success criteria. 
Our work in formalizing the space of communications and building the co-creativity research toolbox will facilitate research in the space of creative AI systems that go beyond algorithmic generative capabilities but also look at how those capabilities support human-agent coordination in creative work.
By doing so, we 
aim to improve the natural collaboration between human creators and AI generation agents, enabling human creators to better achieve their creative intentions.

\bibliography{references}
\end{document}